\documentclass[conference]{IEEEtran}
\ifCLASSINFOpdf
  \usepackage[pdftex]{graphicx}
\else
\fi

\usepackage[cmex10]{amsmath}
\usepackage{amssymb}

\usepackage[lined,algonl,boxed]{algorithm2e}

\usepackage{stfloats}

\usepackage{citesort}

\begin{document}
%
\title{Enhanced Image Classification With a Fast-Learning Shallow Convolutional Neural Network}

\author{\IEEEauthorblockN{Mark D. McDonnell and Tony Vladusich}
\IEEEauthorblockA{Computational and Theoretical Neuroscience Laboratory, Institute for Telecommunications Research,\\School of Information Technology and Mathematical Sciences,\\
University of South Australia\\
Mawson Lakes, SA 5095, Australia\\
Email: mark.mcdonnell@unisa.edu.au}
}

\maketitle

\begin{abstract}
We present a neural network architecture and training method designed to enable very rapid training and low implementation complexity. Due to its training speed and very few  tunable parameters, the method has strong potential for applications requiring frequent retraining or online training. The approach is characterized by (a) convolutional filters based on biologically inspired visual processing filters, (b) randomly-valued classifier-stage input weights, (c) use of least squares regression to train the classifier output weights in a single batch, and (d) linear classifier-stage output units.  We demonstrate the efficacy of the method by applying it to image classification. Our results match existing state-of-the-art results on the MNIST (0.37\% error) and NORB-small (2.2\% error) image classification databases, but with very fast training times compared to standard deep network approaches. The network's performance on the Google Street View House Number (SVHN) (4\% error) database is  also competitive with state-of-the art methods.

\end{abstract}

\IEEEpeerreviewmaketitle

\section{Introduction}

State-of-the-art performance on many image classification databases has been achieved recently using multilayered (i.e., {\em deep}) neural networks~\cite{Schmidhuber.15}.
Such performance generally relies on a convolutional feature extraction stage to obtain invariance to translations, rotations and scale~\cite{LeCun.98,Coates.11,Coates.11a,Le.10}. Training of deep networks, however, often requires significant resources, in terms of time, memory and computing power (e.g. in the order of hours on GPU clusters). Tasks that require online learning, or periodic replacement of all network weights based on fresh data may thus not be able to benefit from deep learning techniques. It is desirable, therefore, to seek very rapid training methods, even if this is potentially at the expense of a small performance decrease. 

Recent work has shown that good performance on image classification tasks can be achieved in `shallow' convolutional networks---neural architectures containing a single training layer---provided sufficiently many features are extracted~\cite{Coates.11}. Perhaps surprisingly, such performance arises even with the use of entirely random convolutional filters or filters based on randomly selected patches from training images~\cite{Coates.11a}. Although application of a relatively large numbers of filters is common (followed by spatial image smoothing and downsampling), good classification performance can also be obtained with a sparse feature representation (i.e. relatively few filters and minimal downsampling)~\cite{Le.10}.

Based on these insights and the goal of devising a fast training method, we introduce a method for combining several existing general techniques into what is equivalent to a five layer neural network (see Figure~\ref{fig_sim}) with only a single trained layer (the output layer), and show that the method:
\begin{enumerate}
\item produces state-of-the-art results on  well known image classification databases;
\item is trainable in times in the order of minutes (up to several hours for large training sets) on standard desktop/laptop  computers;
\item is sufficiently versatile that the same hyper-parameter sets can be applied to different datasets and still produce results comparable to dataset-specific optimisation of hyper-parameters.
\end{enumerate}

The fast training method  we use has been developed independently several times~\cite{Schmidt.92,Chen.96,Eliasmith,Huang.04} and has gained increasing recognition in recent years---see~\cite{Eliasmith.12,Stewart.14a,Huang.12,Huang.14} for recent reviews of the different contexts and applications. The network architecture  in the classification stage is that of a three layer neural network comprised from an input layer, a hidden layer of nonlinear units, and a linear output layer. The input weights are randomly chosen and untrained, and the output weights are trained in a single batch using least squares regression. Due to the convexity of the objective function, this method ensures the output weights are optimally chosen for a given set of random input weights. The rapid speed of training is due to the fact that the least squares optimisation problem an be solved using an O($KM^2$) algorithm, where $M$ is the number of hidden units and $K$ the number of training points~\cite{McDonnell.15PLOS}. 

When applied  to pixel-level features, these networks can be trained as discriminative classifiers and produce excellent results on simple image databases~\cite{vanSchaik.14,Tapson.14,Yu.12,Zhu.14,McDonnell.15PLOS,Zhu.15} but poor performance on more difficult ones.  To our knowledge, however, the method has not yet been applied to  convolutional features.

Therefore, we  have devised a network architecture (see Figure~\ref{fig_sim}) that consists of three key elements that work together to ensure fast learning and good classification performance: namely, the use of (a) convolutional feature extraction, (b) random-valued input weights for classification, (c) least squares training of output weights that feed in to (d) linear output units. We apply our network to several  image classification databases, including MNIST~\cite{MNIST}, CIFAR-10~\cite{Krizhevsky}, Google Street View House Numbers (SVHN)~\cite{SVHN} and NORB~\cite{LeCun.04}. The network produces state-of-the-art classification results on MNIST and NORB-small databases and near state-of-the-art performance on SVHN. 

These promising results are presented in this paper to demonstrate the potential benefits of the method; clearly further innovations within the method are required if it is to be competitive on harder datasets like CIFAR-10, or Imagenet. We expect that the most likely avenues for improving our presented results for CIFAR-10, whilst retaining the method's core attributes, are (1) to introduce limited training of the Stage 1 filters by generalizing the method of~\cite{Yu.12}; (2)  introduction of training data augmentation. We aim to pursuing these directions in our future work.

The remainder of the paper is organized as follows. Section~\ref{S:2} contains a generic description of the network architecture and the algorithms we use for obtaining convolutional features and classifying inputs based on them. Section~\ref{S:3} describes how the generic architecture and training algorithms are {\em specifically} applied to four well-known benchmark image classification datasets. Next, Section~\ref{S:4} describes the results we obtained for these datasets, and finally the paper concludes with discussion and remarks in Section~\ref{S:5}.

%


\section{Network architecture and training algorithms}\label{S:2}

\begin{figure*}[!ht]
\centering
\includegraphics[width=2\columnwidth]{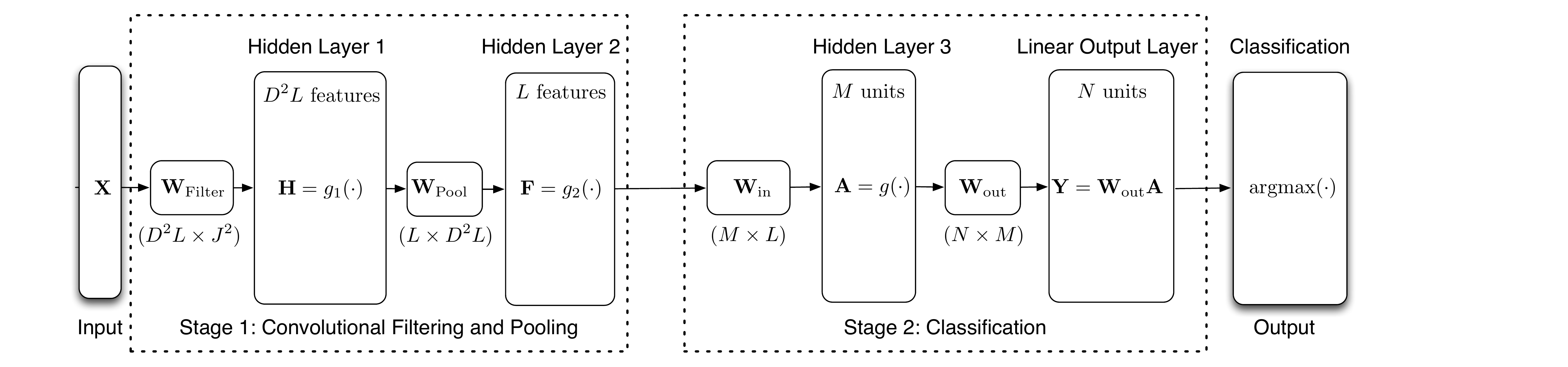}
 \caption{Overall network architecture. In total there are three hidden layers, plus an input layer and a linear output layer. There are two main stages: a convolutional filtering and pooling stage, and a classification stage. Only the final layer of weights, ${\bf W}_{\rm out}$ is learnt, and this is achieved in a single batch using least squares regression. Of the remaining weights matrices, ${\bf W}_{\rm Filter}$ is specified and remains fixed, e.g. taken from Overfeat~\cite{Sermanet.14_overfeat}; ${\bf W}_{\rm  Pool}$ describes standard average pooling and downsampling; and ${\bf W}_{\rm in}$ is set randomly or by using the method of~\cite{Zhu.15} that specifies the weights by sampling examples of the training distribution, as described in the text. Other variables shown are as follows: $J^2$ is the number of pixels in an image, $L$ is the number of features extracted per image, $D$ is a downsampling factor,  $M$ is the number of hidden units in the classifier stage and $N$ is the number of classes. }
\label{fig_sim}
\end{figure*}

The overall network is shown in Figure~\ref{fig_sim}. There are three hidden layers  with nonlinear units, and four layers of weights. The first layer of weights is the convolutional filter layer. The second layer is a pooling (low pass filtering) and downsampling layer. The third layer is a random projection layer. The fourth layer is the only trained layer. The output layer has linear units.

The network can be conceptually divided into two stages and two algorithms, that to our knowledge have not previously been combined. The first stage is the convolutional feature extraction stage, and largely follows that of existing approaches to image classification~\cite{Coates.11,Coates.11a,Le.10,Sermanet.12}. The second stage is the classifier stage, and largely follows the approach of~\cite{Zhu.15,McDonnell.15PLOS}.  We now describe the two stages in detail.

\subsection{Stage 1 Architecture: Convolutional filtering and pooling}\label{S:Stage1}

The algorithm we apply to extract features from images (including those with multiple channels) is summarised in {\bf Algorithm~\ref{algorithm1}}. Note that the details of the filters ${\bf h}_{i,c}$ and ${\bf h}_{\rm p}$ described in  {\bf Algorithm~\ref{algorithm1}} are given in Section~\ref{S:filters}, but here we introduce the size of these two-dimensional filters as $W\times W$ and $Q\times Q$. The functions $g_1(\cdot)$ and $g_2(\cdot)$ are nonlinear transformations applied termwise to matrix inputs to produce matrix outputs of the same size. The symbol * represents two-dimensional convolution.

\begin{algorithm}
\SetKwInOut{Input}{Input}
\SetKwInOut{Output}{Output}
\caption{Convolutional feature detection.}
\Input{Set of $K$ images, ${\bf x}_{k}$, each with $C$ channels}
\Output{Feature vectors, ${\bf f}_{k}$, $k=1,\dots K$}
\BlankLine
\ForEach{${\bf x}_{k}$}{
split ${\bf x}_{k}$ into channels, ${\bf x}_{k,c}$, $c=1,\dots C$\\
\ForEach{$i=1,\dots P$ {\rm filters}}{
Apply filter to each channel: ${\bf y}_{i,k,c} \leftarrow {\bf h}_{i,c}*{\bf x}_{k,c}$\\
Apply termwise nonlinearity: ${\bf z}_{i,k,c} \leftarrow g_1({\bf y}_{i,k,c})$\\
Apply lowpass filter: ${\bf w}_{i,k,c} \leftarrow {\bf h}_{\rm p}*{\bf z}_{i,k,c}$\\
Apply termwise nonlinearity: ${\bf s}_{i,k,c} \leftarrow g_2({\bf w}_{i,k,c})$\\
Downsample: $\hat{{\bf s}}_{i,k,c} \leftarrow {{\bf s}}_{i,k,c}$\\
Concatenate  channels:  ${\bf r}_{i,k} \leftarrow \left[\hat{{\bf s}}_{i,k,1}|\dots |\hat{{\bf s}}_{i,k,C}\right]$\\
Normalize: ${\bf f}_{i,k} = {{{\bf r}}_{i,k}}/({{\bf r}_{i,k}{\bf 1}^{\top}})$
}
Concatenate over filters:  ${\bf f}_{k} \leftarrow\left[{{\bf f}}_{1,k}|{{\bf f}}_{2,k}|\dots |{{\bf f}}_{P,k}\right]$\\
}
\label{algorithm1}
\end{algorithm}

This sequence of steps in  {\bf Algorithm~\ref{algorithm1}} suggest looping over all images and channels sequentially. However, the following mathematical formulation of the  algorithm indicates a standard layered neural network formulation of this algorithm is applicable, as shown in Figure~\ref{fig_sim}, and therefore that computation of all features (${\bf f}_{k},k=1,..K$ ) can be obtained in one shot from a $K$-column matrix containing a batch of $K$ training points.

The key to this formulation is to note  that since convolution is a linear operator, a matrix can be constructed that when multiplied by a data matrix produces the same result as convolution applied to one instance of the data. Hence, for a total of $L$ features per image, we introduce the following matrices. 

Let 
\begin{itemize}
\item ${\bf F}$ be a feature matrix of size $L\times K$;
\item ${\bf X}$ be a data matrix with $K$ columns;
\item ${\bf W}_{\rm Filter}$ be a concatenation of the $CP$ convolution matrices corresponding to ${\bf h}_{i,c}~i=1,\dots P,c=1,\dots C$;
\item ${\bf W}_0$ be a convolution matrix corresponding to ${\bf h}_l$, that also down samples by a factor of $D$;
\item ${\bf W}_{\rm Pool}$ be a block diagonal matrix containing $CP$ copies of ${\bf W}_0$ on the diagonals.
\end{itemize}
The entire flow described in Algorithm~\ref{algorithm1} can  be written mathematically as
\begin{equation}
{\bf F} = g_2({\bf W}_{\rm Pool}~g_1({\bf W}_{\rm Filter}{\bf X})),
\end{equation}
where  $g_1(\cdot)$ and $g_2(\cdot)$ are applied term by term to all elements of their arguments. The matrices  ${\bf W}_{\rm Filter}$ and ${\bf W}_{\rm Pool}$ are sparse Toeplitz matrices. In practice we would not form them directly, but instead form one pooling matrix, and one filtering matrix for each filter, and sequential apply each filter to the entire data matrix, ${\bf X}$.

We use a particular  form for the nonlinear hidden-unit functions $g_1(\cdot)$ and $g_2(\cdot)$  inspired by {\em LP-pooling}~\cite{Sermanet.12}, which is of the form $g_1(u) = u^p$ and
$g_2(v) = v^{\frac{1}{p}}$. For example, with $p=2$ we have
\begin{equation}
{\bf F} = \sqrt{{\bf W}_{\rm Pool}~({\bf W}_{\rm Filter}{\bf X})^2}.
\end{equation}

An intuitive explanation for the use of LP-pooling is as follows. First, note that each hidden unit receives as input a linear combination of a patch of the input data, i.e. $u$ in $g_1(u)$ has the form $u=\sum_{j=1}^{W^2}h_{i,c,j}x_{i,j}$. Hence, squaring $u$ results in a sum that contains  terms proportional to $x_{i,j}^2$ and terms proportional to products of each $x_{i,j}$. Thus, squaring is a simple way to produce hidden layer responses that depend on the product of pairs of input data elements, i.e. {\em interaction terms}, and this is important for discriminability.  Second, the square root transforms the distribution of the hidden-unit responses; we have observed that in practice, the result of the square root operation is often a distribution that is closer to Gaussian than without it, which helps to regularise the least squares regression method of training the output weights.

However, as will be described shortly, the classifier of Stage 2 also has a square nonlinearity. Using this nonlinearity, we have found that classification performance is generally optimised by taking the square root of the input to the random projection layer.  Based on this observation, we do not strictly use LP-pooling, and instead set 
\begin{equation}\label{g1_a}
g_1(u) = u^2,
\end{equation}
and
\begin{equation}\label{g2_a}
g_2(v) = v^{0.25}.
\end{equation}
This effectively combines the implementation of L2-pooling, and the subsequent square root operation.

\subsection{Stage 2 Architecture: Classifier}

The following descriptions are applicable whether or not  raw pixels are treated as features or the input is the features extracted in stage 1.
First, we introduce notation. Let:
\begin{itemize}
\item ${\bf F}_{\rm train}$, of size $L\times K$, contain each length $L$ feature vector;
\item ${\bf Y}_{\rm label}$ be an indicator matrix of size $N\times K$, which numerically represents the labels of each training vector, where there are $N$ classes---we set each column to have a $1$ in a single row, corresponding to the label class for each training vector, and all other entries to be zero;
\item ${\bf W}_{\rm in}$, of size $M\times L$ be the real-valued input weights matrix for the classifier stage;
\item ${\bf W}_{\rm out}$, of size $N\times M$ be the real-valued output weights matrix for the classifier stage;
\item the function $g(\cdot)$ be the activation function of each hidden-unit; for example, $g(\cdot)$ may be the logistic sigmoid, $g(z) = 1/(1+\exp(-z))$, or a squarer, $g(z)=z^2$;
\item ${\bf A}_{\rm train}=g({\bf W_{\rm in}}{\bf F_{\rm train}})$, of size $M\times K$, contain the hidden-unit activations that occur due to each feature vector; $g(\cdot)$ is applied termwise to each element in the matrix  ${\bf W_{\rm in}}{\bf F_{\rm train}}$.
\end{itemize}

\subsection{Stage 1 Training: Filters and Pooling}

In this paper we do not employ any form of training for the filters and pooling matrices. The details of the filter weights and form of pooling used for the example classification problems presented in this paper are given Section~\ref{S:3}.

\subsection{Stage 2 Training: Classifier Weights}

The  training approach for the classifier is that described by  e.g.~\cite{Schmidt.92,Chen.96,Eliasmith,Huang.14}. The default situation for these methods is that the input weights, ${\bf W}_{\rm in}$, are generated randomly from a specific distribution, e.g. standard Gaussian, uniform, or bipolar. However, it is known that setting these weights non-randomly based on the training data leads to superior performance~\cite{McDonnell.15PLOS,Tapson.14,Zhu.15}. In this paper, we use the method of~\cite{Zhu.15}.  The input weights can also be trained iteratively, if desired, using single-batch  backpropagation~\cite{Yu.12}.

Given a choice of ${\bf W}_{\rm in}$, the output weights matrix is determined according to
\begin{align}\label{T2}
{\bf W}_{\rm out} = {\bf Y}_{\rm label}{\bf A}_{\rm train}^+,
\end{align}
where ${\bf A}_{\rm train}^+$ is the size $K\times M$ Moore-Penrose pseudo inverse corresponding to ${\bf A}_{\rm train}$. This solution is equivalent to least squares regression applied to an overcomplete set of linear equations, with an $N$-dimensional target.  It is known to often be useful to regularise such problems, and instead solve the following {\em ridge regression} problem~\cite{Huang.12,Huang.14}:
\begin{align}\label{Q}
{\bf W}_{\rm out} &= {\bf Y}_{\rm label}{\bf A}_{\rm train}^{\top}({\bf A}_{\rm train}{\bf A}_{\rm train}^{\top}+c{\bf I})^{-1},
\end{align}
where $c$ is a hyper-parameter and ${\bf I}$ is the $M\times M$ identity matrix. In practice, it is efficient to avoid explicit calculation of the inverse in Equation~(\ref{Q})~\cite{McDonnell.15PLOS} and instead use QR factorisation to solve the following set of $NM$ linear equations for the $NM$ unknown variables in ${\bf W}_{\rm out}$:
\begin{align}\label{Final}
{\bf Y}_{\rm label}{\bf A}_{\rm train}^{\top} &= {\bf W}_{\rm out}({\bf A}_{\rm train}{\bf A}_{\rm train}^{\top}+c{\bf I}).
\end{align}
Above we mentioned two algorithms, and {\bf Algorithm 2} is simply to form ${\bf A}_{\rm train}$ and solve Eqn.~(\ref{Final}), followed by optimisation of $c$ using ridge regression. For large $M$ and $K>M$ (which is typically valid) the runtime bottleneck for this method is typically the O$(KM^2)$ matrix multiplication required to obtain the Gram matrix, ${\bf A}_{\rm train}{\bf A}_{\rm train}^{\top}$.

\subsection{Application to Test Data}

For a total of $K_{\rm test}$ test images contained in a matrix ${\bf X}_{\rm test}$, we first obtain a matrix  ${\bf F}_{\rm test}=g_2({\bf W}_{\rm Pool}~g_1({\bf W}_{\rm Filter}{\bf X_{\rm test}}))$, of size $L\times K_{\rm test}$, by following {\bf Algorithm~\ref{algorithm1}}. The output of the classifier is then the $N\times K_{\rm test}$ matrix
\begin{align}\label{Y_test}
{\bf Y}_{\rm test}& = {\bf W}_{\rm out}~g({\bf W_{\rm in}}{\bf F_{\rm test}})\\
&= {\bf W}_{\rm out}~g({\bf W_{\rm in}}~g_2({\bf W}_{\rm Pool}~g_1({\bf W}_{\rm Filter}{\bf X}))).
\end{align}
Note that we can write the response to all test images in terms of the training data:
\begin{align}
{\bf Y}_{\rm test}& = {\bf Y}_{\rm label}\left(g({\bf W_{\rm in}}{\bf F_{\rm train}})\right)^+~g({\bf W_{\rm in}}{\bf F_{\rm test}})\label{Y_test1a}\\
{\rm~where~~~~~}&\notag\\
{\bf F}_{\rm train}&=g_2({\bf W}_{\rm Pool}~g_1({\bf W}_{\rm Filter}{\bf X_{\rm train}}))\label{Y_test1b}\\
{\bf F}_{\rm test}&=g_2({\bf W}_{\rm Pool}~g_1({\bf W}_{\rm Filter}{\bf X_{\rm test}}))\label{Y_test1c}.
\end{align}
Thus, since the pseudo-inverse, $(\cdot)^+$, can be obtained from Equation~(\ref{Q}), Equations~(\ref{Y_test1a}),~(\ref{Y_test1b}) and~(\ref{Y_test1c}) constitute a closed-form solution for the entire test-data classification output, given specified matrices, ${\bf W}_{\rm filter}$, ${\bf W}_{\rm pool}$ and ${\bf W}_{\rm in}$, and hidden-unit activation functions, $g_1, g_2$, and $g$.

The final classification decision for each image is obtained by taking the index of the maximum value of each column of ${\bf Y}_{\rm test}$.

\section{Image Classification Experiments: Specific Design}\label{S:3}

We  examined the method's performance  when used as a classifier of images. Table~\ref{Table1} lists the attributes of  four well known databases we used. For the two databases comprised from RGB images, we used $C=4$ channels, namely the  raw RGB channels, and a conversion to greyscale. This approach was shown to be effective for SVHN in~\cite{Sermanet}. 

\begin{table}[!ht]
{\footnotesize
\begin{tabular}{|c|c|c|c|c|c|}
\hline
Database & Classes & Training & Test  & Channels & Pixels\\
\hline
MNIST~\cite{MNIST} & 10 & 60000 & 10000 & 1 & 28$\times$28\\
NORB-small~\cite{LeCun.04}\ & 5 & 24300 & 24300 & 2 (stereo) & 32$\times$32\\
SVHN~\cite{SVHN}  & 10 & 604308 & 26032 &  3 (RGB) & 32$\times$32\\
CIFAR-10~\cite{Krizhevsky} & 10 & 50000 & 10000 & 3 (RGB) & 32$\times$32\\
\hline
\end{tabular}
~\\
\caption{\bf{Image Databases. Note that the NORB-small database consists of images of size $96 \times 96$ pixels, but we first downsampled all training and test images to $32 \times 32$ pixels, as in~\cite{Le.10}.}}\label{Table1}
}
\end{table}%

\subsection{Preprocessing}

All raw image pixel values were scaled to the interval $[0,1]$. Due to the use of quadratic nonlinearities and LP-pooling, this scaling does not affect performance. The only other preprocessing done was as follows:
\begin{enumerate}
\item MNIST: None;
\item NORB-small: downsample from 96$\times$96 to 32$\times$32, for implementation efficiency reasons (this is consistent with some previous work on NORB-small, e.g.~\cite{Le.10});
\item SVHN: convert from 3 channels to 4 by adding a conversion to greyscale from the raw RGB. We found that local and/or global contrast enhancement only diminished performance;
\item CIFAR-10: convert from 3 channels to 4 by adding a conversion to greyscale from the raw RGB; apply ZCA whitening to each channel of each image, as in~\cite{Coates.11}.
\end{enumerate}

\subsection{Stage 1 Design: Filters and Pooling}\label{S:s1design}\label{S:filters}

Since our objective here was to train only a single layer of the network, we did not seek to train the network to find filters optimised for the training set. Instead, for the size $W\times W$ two-dimension filters, ${\bf h}_{i,c}$, we considered the following options:
\begin{enumerate}
\item simple rotated bar and corner filters, and square uniform centre-surround filters;
\item filters trained on Imagenet and made available in Overfeat~\cite{Sermanet.14_overfeat}; we used only the 96 stage-1 `accurate' 7$\times$7 filters;
\item patches obtained  from the central $W\times W$ region of randomly selected training images, with $P/N$ training images from each class.
\end{enumerate}
The  filters from Overfeat\footnote{Available from http://cilvr.nyu.edu/doku.php?id=software:overfeat:start} are RGB filters. Hence, for the databases with RGB images, we applied each channel of the filter to the corresponding channel of each image. When applied to greyscale channels, we converted the Overfeat filter to greyscale. For NORB, we applied the same filter to both stereo channels. For all filters, we subtract the mean value over all $W^2$ dimensions in each channel, in order to ensure a mean of zero in each channel.

In implementing the two-dimensional convolution operation required for filtering the raw images using ${\bf h}_{i,c}$, we obtained only the central `valid' region, i.e.~for images of size $J\times J$, the total dimension of the valid region is $(J-W+1)^2$. Consequently, the total number of features per image obtained prior to pooling, from $P$ filters, and images with $C$ channels is $L= CP(J-W+1)^2$.

In previous work, e.g.~\cite{Sermanet.12}, the form of the $Q\times Q$ two-dimension filter, ${\bf h}_{\rm p}$ is a normalised Gaussian. Instead, we used a simple summing filter, equivalent to a kernel with all entries equal to the same value, i.e.
\begin{equation}\label{h_pool}
{\bf h}_{{\rm p},u,v}=\frac{1}{Q^2},\quad~u=1,\dots Q, v=1,\dots Q.
\end{equation}
In implementing the two-dimensional convolution operation required for filtering using ${\bf h}_{{\rm p}}$, we obtained the `full' convolutional region, which for images of size $J\times J$ is $(J-W+Q)^2$, given the `valid' convolution first applied using ${\bf h}_{i,c}$, as described above.

The remaining part of the pooling step is to downsample each image dimension by a factor of $D$, resulting in a total of $\hat{L} = L/D^2$ features per image. In choosing $D$, we experimented with a variety of scales before settling on the value  shown in Table~\ref{Table2}. We note there exists an interesting tradeoff between the number of filters $P$, and the downsampling factor, $D$. For example, in~\cite{Coates.11}, $D=L/2$, whereas in~\cite{Le.10} $D=1$.  We found that, up to a point, smaller $D$ enables a smaller number of filters, $P$, for comparable performance.

The hyper-parameters we used for each dataset are shown in Table~\ref{Table2}.

\begin{table}[!ht]
{\footnotesize
\begin{tabular}{|c|c|c|c|c|}
\hline
Hyper-parameter & MNIST & NORB & SVHN & CIFAR-10\\
\hline
Filter size, $W$ & 7 & 7 & 7 & 7\\
Pooling size, $Q$ & 8 & 10 & 7 & 7\\
Downsample factor,  $D$ & 2 & 2 & 5 & 3\\
\hline
\end{tabular}
~\\
\caption{\bf{Stage 1 Hyper-parameters (Convolutional Feature Extraction).  }}\label{Table2}
}
\end{table}%

\subsection{Stage 2 Design: Classifier projection weights}

To construct the  matrix ${\bf W}_{\rm in}$ we use the method proposed by~\cite{Zhu.14}. In this method, each row of the matrix ${\bf W}_{\rm in}$ is chosen to be a normalized difference between the data vectors corresponding to randomly chosen examples from distinct classes of the training set. This method has previously been shown to be superior to setting the weights to values chosen from random distributions~\cite{Zhu.14,McDonnell.15PLOS}.

For the nonlinearity in the classifier stage hidden units, $g(z)$, the typical choice in other work~\cite{Huang.14}  is  a sigmoid. However, we found it sufficient (and much faster in an implementation) to use the quadratic nonlinearity. This suggests that good image classification is strongly dependent on  the presence of interaction terms---see the discussion about this in Section~\ref{S:Stage1}.

\subsection{Stage 2 Design: Ridge Regression parameter}

With these choices, there remains only two hyper-parameters for the Classifier stage: the regression parameter, $c$, and the number of hidden-units, $M$. In our experiments, we examined classification error rates as a function of varying $M$. For each $M$, we can optimize $c$ using cross-validation. However, we also found that a good generic heuristic for setting $c$ was
\begin{equation}
c = \frac{N^2}{M^2}{\rm min}({\rm diag}({\bf A}_{\rm train}{\bf A}_{\rm train}^{\top})),
\end{equation} 
and this reduces the number of hyper-parameters for the classification stage to just one: the number of hidden-units, $M$.

\subsection{Stage 1 and 2 Design: Nonlinearities}

For the hidden-layer nonlinearities, to reiterate, we use:
\begin{equation}\label{nonlin}
g_1(u) = u^2,~g_2(v) = v^{0.25},~g(z) = z^2.
\end{equation}

\section{Results}\label{S:4}

We examined the performance of the network on classifying the test images in the four chosen databases, as a function of the number of filters, $P$, the downsampling rate $D$, and the number of hidden units in the classifier stage, $M$. We use the maximum number of channels, $C$, available in each dataset (recall from above that we convert RGB images to greyscale, as a fourth channel). 

We considered the three kinds of untuned filters described in Section~\ref{S:s1design}, as well as combinations of them. We did not exhaustively consider all options, but settled on the Overfeat filters as being marginally superior for NORB, SVHN and CIFAR-10 (in the order of 1\% in comparison with other options), while hand-designed filters were superior for MNIST, but only marginally compared to randomly selected patches from the training data. There is clearly more that can be investigated to determine whether hand-designed filters can match trained filters when using the method of this paper.

\subsection{Summary of best performance attained}

The best performance we achieved is summarised in Table~\ref{Table4}. 

\begin{table}[!ht]
{\footnotesize
\begin{tabular}{|c|c|c|c|c|c|}
\hline
Database  & $C$ & $M$ & $P$& Our best & State-of-the-art\\
\hline
MNIST & 1& 12000  & 60 & 0.37\% & 0.39\%~\cite{Mairal.14,Lee.14}\\
NORB-small & 2&  3200 & $60$ & 2.21\% & 2.53\%~\cite{Ciresan.11}\\
SVHN  &  4& 40000 & $96$ & 3.96\%& 1.92\%~\cite{Lee.14}\\
CIFAR-10 &   4& 40000 & $96$ & 24.14\% & 9.78\%~\cite{Lee.14}\\ 
\hline
\end{tabular}
~\\
\caption{\bf{Results for various databases. The state-of-the-art result listed for MNIST and CIFAR-10 can be improved by augmenting the training set with distortions and other methods~\cite{Simard.03,Ciresan.10,Ciresan.12}; we have not done so here, and report state-of-the-art only for methods not doing so.}}\label{Table4}
}
\end{table}%

\subsection{Trend with increasing $M$}

\begin{figure}[!ht]
\centering
\includegraphics[width=0.9\columnwidth]{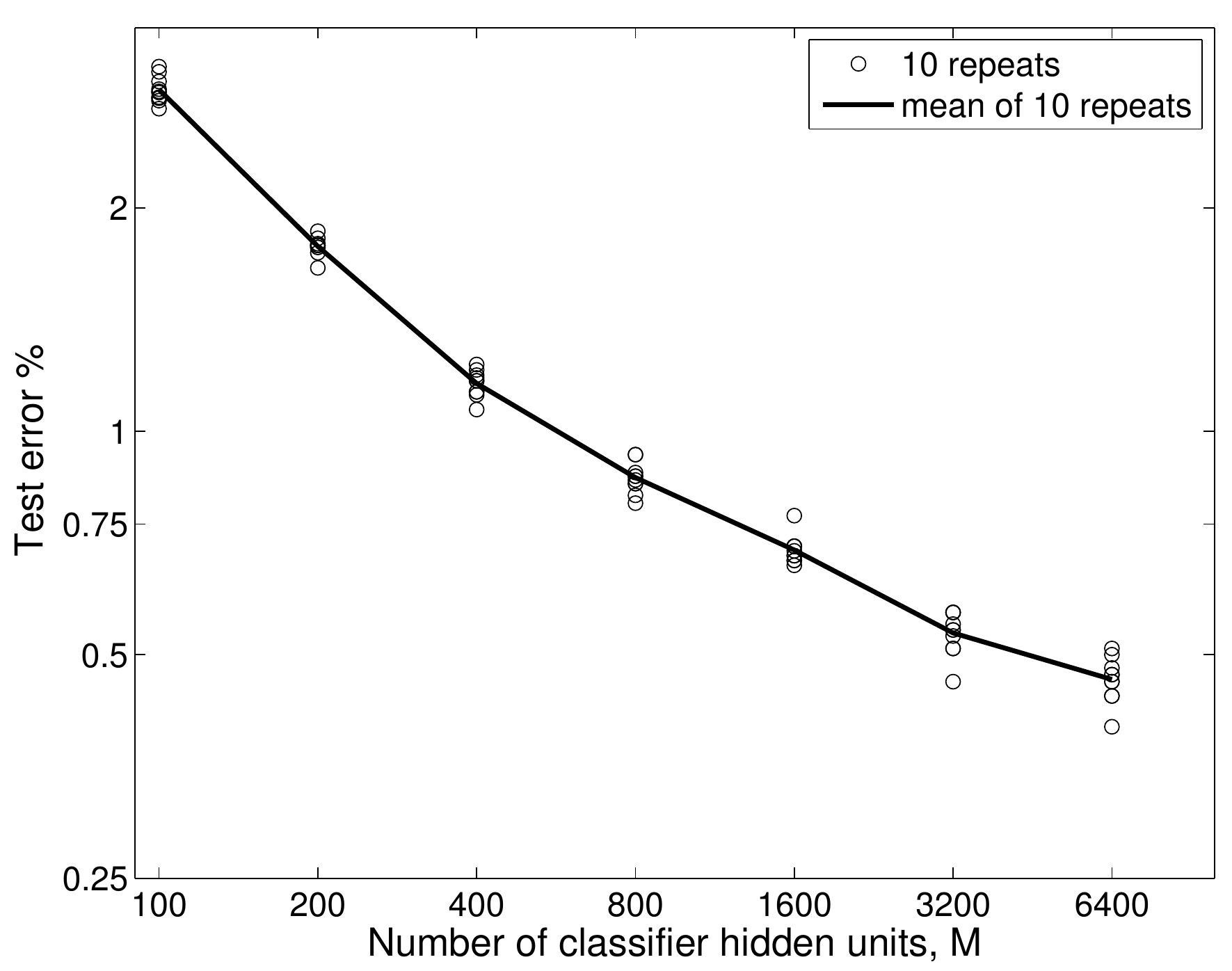}
 \caption{Example set of error percentage value on the 10000 MNIST test images, for ten repetitions of the selection ${\bf W}_{\rm in}$. The best result shown is 40 errors out of 10000. Increasing $M$ above 6400 saturates in performance.}
\label{fig_MNIST}
\end{figure}

We now use MNIST as an example to indicate how classification performance scales with the number of hidden units in the classifier stage, $M$. The remain parameters were $W=7$, $D=3$ and $P=43$, which included hand-designed filters comprised from 20 rotated bars (width of one pixel), 20 rotated corners (dimension 4 pixels) and 3 centred squares (dimensions 3, 4 and 5 pixels), all with  zero mean. The rotations were of binary filters and used standard pixel value interpolation. Figure~\ref{fig_MNIST} shows a power law-like decrease in error rate as $M$ increases, with a linear trend on the log-log axes. The best error rate shown on this figure is 0.40\%. As shown in Table~\ref{Table4}, we have attained a best repeatable rate of $0.37$\% using 60 filters and $D=2$. When we combined Overfeat filters with hand-designed filters and randomly selected patches from the training data, we obtained up to $0.32$\% error on MNIST, but this was an outlier since it was not repeatedly obtained by different samples of ${\bf W}_{\rm in}$.

\subsection{Indicative training times}

For an implementation in Matlab on a PC with 4 cores and 32 GB of RAM, for MNIST (60000 training points) the total time required to generate all features for all 60000 training images from one filter is approximately 2 seconds. The largest number of filters we used to date was 384 (96 RGB+greyscale), and when applied to SVHN ($\sim$600000 training points), the total run time for feature extraction is then about two hours (in this case we used batches of size 100000 images). 

The runtime we achieve for feature generation benefits from carrying out convolutions using matrix multiplication applied to large batches simultaneously; if instead we iterate over all training images individually, but still carry out convolutions using matrix multiplication, the time for generating features approximately doubles.   Note also that we employ Matlab's sparse matrix data structure functionality to represent ${\bf W}_{\rm Filter}$ and ${\bf W}_{\rm Pool}$, which also provides a speed boost when multiplying these matrices to carry out the convolutions.  If we do not use the matrix-multiplication method for convolution, and instead apply two-dimensional convolutions to each individual image, the feature generation is slowed even more.

For the classifier stage, on MNIST with $M=6400$, the runtime is approximately 150 seconds for $D=3$ (there is a small time penalty for smaller $D$, due to the larger dimension of the input to the classifier stage). Hence, the total run time for MNIST with 40 filters and $M=6400$ is in the order of 4 minutes to achieve a correct classification rate above  99.5\%. With fewer filters and smaller $M$, it is simple to achieve over 99.2\% in a minute or less.

For SVHN and CIFAR-10 where we scaled up to $M=40000$, the run time bottleneck is the classifier, due to the $O(KM^2)$ runtime complexity. We found it necessary to use a PC with more RAM (peak usage was approximately 70 GB) for $M>20000$. In the case of $M=40000$, the network was trained in under an hour on CIFAR-10, while SVHN took about 8-9 hours.  Results within a few percent of our best, however, can be obtained in far less time.

\section{Discussion and Conclusions}\label{S:5}

As stated in the introduction, the purpose of this paper is to highlight the potential benefits of the method presented, namely that it can attain excellent results with a rapid training speed  and low implementation complexity, whilst only suffering from reduced performance relative to state-of-the-art on particularly hard problems.

In terms of efficacy on classification tasks, as shown in Table~\ref{Table4}, our best result (0.37\% error rate) surpasses the best ever reported performance for classification of the MNIST test set when no  augmentation of the training set is done. We have also achieved, to our knowledge, the best performance reported in the literature for the NORB-small database, surpassing the previous best~\cite{Ciresan.11} by about 0.3\%.

For SVHN, our best result is within $\sim2$\% of state-of-the-art. It is highly likely that using  filters trained on the SVHN database rather than on Imagenet would reduce this gap, given the structured nature of digits, as opposed to the more complex nature of Imagenet images. Another avenue for closing the gap on state-of-the-art using the same filters would be to increase $M$ and decrease $D$, thus resulting in more features and more classifier hidden units. Although we increased $M$ to 40000, we did not observe saturation in the error rate as we increased $M$ to this point.

For CIFAR-10, it is less clear what is lacking in our method in comparison with the gap of about 14\% to state-of-the-art methods. We note that CIFAR-10 has relatively few training points, and we observed that the gap between classification performance on the actual training set, in comparison with the test set, can be up to 20\%. This suggests that designing enhanced methods of regularisation (e.g. methods similar to dropout in the convolutional stage, or data augmentation) are necessary to ensure our method can achieve good performance on  CIFAR-10.  Another possibility is to use a nonlinearity in the classifier stage that ensures the hidden-layer responses reflect higher order correlations than possible from the squaring function we used. However, we expect that training the convolutional filters in Stage 1 so that they extract features that are more discriminative for the specific dataset will be the most likely enhancement for improving results on CIFAR-10.

Finally, we note that there exist iterative approaches for training the classifier component of Stage 2 using least squares regression, and without training the input weights---see, e.g.,~\cite{Tapson.13,Widrow.13,McDonnell.15PLOS}. These methods can be easily adapted for use with the convolutional front-end, if, for example,  additional batches of training data become available, or if the problem involves online learning. 

In closing, following acceptance of this paper, we became aware of a newly published paper that combines convolutional feature extraction with least squares regression training of classifier weights to obtain good results for the NORB dataset~\cite{Huang.15}. The three main differences between the method of the current paper and the method of~\cite{Huang.15} are as follows. First, we used a hidden layer in our classifier stage, whereas~\cite{Huang.15} solves for output weights using least squares regression  applied to  the output of the pooling stage. Second, we used a variety of methods for the convolutional filter weights, whereas~\cite{Huang.15} uses orthogonalised random weights only.  Third, we downsample following pooling, whereas~\cite{Huang.15} does not do so.

\section*{Acknowledgment}

Mark D. McDonnell's contribution was by supported by an Australian Research Fellowship from the Australian Research Council (project number DP1093425).  We gratefully acknowledge Prof David Kearney and Dr Victor  Stamatescu from University of South Australia and Dr Sebastien Wong of DSTO, Australia, for useful discussions and provision of computing resources. We also acknowledge discussions with Prof Philip De Chazal of University of Sydney.

\end{document}